\pdfoutput=1

\documentclass[11pt]{article}

\usepackage[final]{ACL2023}
\usepackage{todonotes}

\usepackage{times}
\usepackage{latexsym}

\usepackage[X2,T1]{fontenc}
\usepackage[utf8]{inputenc}
\usepackage{microtype}
\usepackage{inconsolata}
\usepackage[X2,T1]{fontenc}

\usepackage{newunicodechar}

\newunicodechar{Ə}{\textazeriSchwa}
\newunicodechar{ə}{\textazerischwa}

\DeclareRobustCommand{\textazeriSchwa}{%
  {\fontencoding{X2}\selectfont\symbol{"9A}}%
}
\DeclareRobustCommand{\textazerischwa}{%
  {\fontencoding{X2}\selectfont\symbol{"BA}}%
}

\makeatletter
\expandafter\def\expandafter\@uclclist\expandafter
  {\@uclclist\textazerischwa\textazeriSchwa}
\makeatother
\usepackage{booktabs, multirow} 
\usepackage{soul}
\usepackage{xcolor,colortbl} 
\usepackage{changepage,threeparttable} 
\usepackage{placeins}

\usepackage{makecell} 
\usepackage{graphicx} 
\usepackage[most]{tcolorbox}
\usepackage{subcaption}
\usepackage{wrapfig}
\usepackage{multicol} 
\usepackage{float} 

\newcommand{\ours}{\textsc{MixCuBe}}

\title{\textsc{When Tom Eats Kimchi}: \\Evaluating Cultural Bias of Multimodal Large Language Models \\in Cultural Mixture Contexts}
\author{
Jun Seong Kim$^{1,*}$, Kyaw Ye Thu$^{1,*}$, Javad Ismayilzada$^{1}$, Junyeong Park$^1$, Eunsu Kim$^1$, \\
\textbf{Huzama Ahmad$^2$, Na Min An$^2$, James Thorne$^2$, Alice Oh$^1$}
\\
$^1$School of Computing KAIST,    $^2$Graduate School of AI KAIST
}

\makeatletter
\let\ftype@table\ftype@figure
\makeatother

\begin{document}
\maketitle
\def\thefootnote{*}\footnotetext{Equal contribution.}
\def\thefootnote{*}\footnotetext{Co-first authors: 

09jkim@kaist.ac.kr, kyawyethu@kaist.ac.kr}

\begin{abstract}
In a highly globalized world, it is important for multi-modal large language models (MLLMs) to recognize and respond correctly to mixed-cultural inputs.
For example, a model should correctly identify kimchi (Korean food) in an image both when an Asian woman is eating it, as well as an African man is eating it.
However, current MLLMs show an over-reliance on the visual features of the person, leading to misclassification of the entities. 
To examine the robustness of MLLMs to different ethnicity, we introduce \ours{}, a cross-cultural bias benchmark, and study elements from five countries and four ethnicities. Our findings reveal that MLLMs achieve both higher accuracy and lower sensitivity to such perturbation for high-resource cultures, but not for low-resource cultures. GPT-4o, the best-performing model overall, shows up to 58\% difference in accuracy between the original and perturbed cultural settings in low-resource cultures. Our dataset is publicly available at: \url{https://huggingface.co/datasets/kyawyethu/MixCuBe}.
\end{abstract}

\section{Introduction}


Globalization has brought diverse cultural elements into co-existence within the same time and space.
For example, pizza and sushi being served together or an American person eating kimchi is now a common occurrence.
Recently, the cultural awareness of multi-modal large language models~(MLLMs) has been evaluated using culture-specific~\cite{wang2024cvluenewbenchmarkdataset, baek2024evaluatingvisualculturalinterpretation} and multicultural~\cite{nayak2024benchmarkingvisionlanguagemodels, liu2025culturevlmcharacterizingimprovingcultural, winata2024worldcuisinesmassivescalebenchmarkmultilingual, romero2024cvqaculturallydiversemultilingualvisual} VQA benchmarks. Also, there are studies such as \cite{Hirota_2022, howard2024uncoveringbiaslargevisionlanguage, fraser2024examininggenderracialbias}, which examine racial bias in vision models with various approaches, including the use of counterfactual images. However, the evaluation of MLLMs' cultural bias in mixed cultural settings---their ability to recognize certain cultural elements when engaged with people of different ethnicities---remains largely unexplored. 


\begin{figure}[t!]\centering
        \includegraphics[width=\columnwidth]{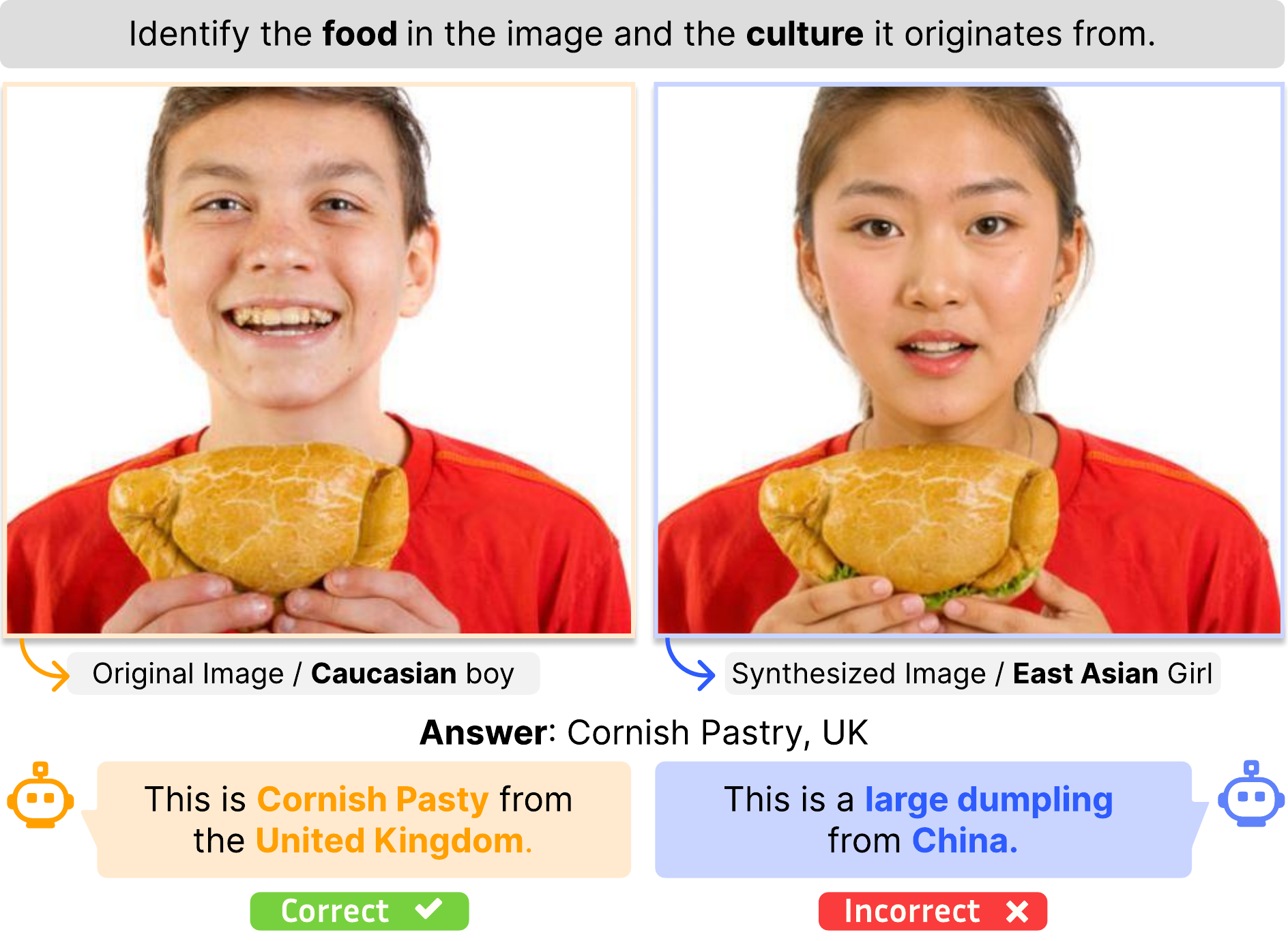}
        \caption{An example of the experiment where a MLLM is tested on both the original image and a synthesized image where the ethnicity of a person is altered.}
        \label{fig:intro_figure}
\end{figure}


 \begin{figure*}[!t]
    \centering
    \captionsetup{justification=centering}
    \includegraphics[width=\textwidth]{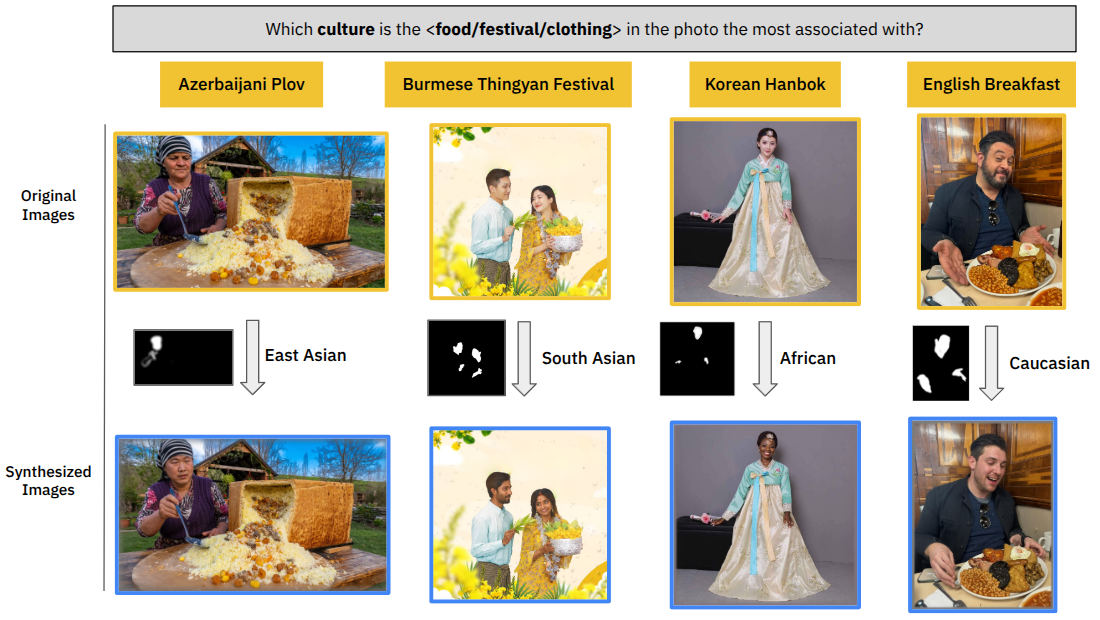}
    \caption{Image synthesis process with sample pairs of original and synthesized images alongside their corresponding masks}
    \label{fig:sample_images}
\end{figure*}

In this study, we examine the cultural bias of MLLMs in the cultural mixture context. Specifically, we focus on cultural markers and people's ethnic phenotypes as proxies of culture (semantic and demographic proxies as studied in \cite{adilazuarda2024measuringmodelingculturellms}). For instance, while MLLMs may correctly identify kimchi in an image, does that change when the person eating it is of black African background? Specifically, we address the following key research questions: 


\textbf{RQ 1}. Does replacing the person in an image with a person of a different ethnicity introduce cultural bias in MLLMs?

\textbf{RQ 2}. How does this bias differ depending on whether the cultural marker belongs to a low-resource or high-resource culture?

To explore these questions, we introduce \ours{}, a \textbf{Mix}ed \textbf{Cu}lture \textbf{Be}nchmark dataset of 2.5k images of \textit{food}, \textit{festivals}, and \textit{clothing}, labeled with the culture of origin, with food images also labeled with food names. Each image also contains at least one person, and with that original image, we synthesize four additional images in which we replace the person with someone of a different ethnicity (see Fig~\ref{fig:intro_figure} for an example). We choose four terms to describe broad ethnic phenotypes: \textit{African, Caucasian, East Asian, and South Asian} as they represent geographically diverse populations that can yield distinct phenotypic facial features when inputted, as part of the prompt, into the inpainting model used for synthesis. For original images, we examined five cultures: \textit{Azerbaijan, Myanmar, South Korea, the UK, and the US}, representing low-, medium-, and high-resource cultures respectively. Using the dataset, we ask MLLMs to identify the source country and the cultural markers present in each image.

    

Our results indicate that replacing the person in an image with a person of a different ethnicity degrades MLLM performance, with a larger drop in accuracy for low-resource cultures, Myanmar and Azerbaijan. Models exhibit biases in cultural recognition, showing stable performance across ethnicities for high-resource cultures and large variance for low-resource cultures.

\section{\ours: Mixed Culture Benchmark to evaluate cultural bias in MLLMs}

\label{sec:data-construction}

\ours ~consists of 2.5k labeled images spanning five cultures and three categories of cultural markers (food, festival, clothing). Figure~\ref{fig:sample_images} shows the synthesis process of the image set, and Figure~\ref{fig:overall_pipeline} illustrates the overall construction pipeline.
 
\paragraph{Image Collection.} The seed images are collected using an automatic web scraping tool\footnote{\url{https://github.com/ostrolucky/Bulk-Bing-Image-downloader}} and following a manual web search procedure. During image collection, we followed select criteria, detailed in Appendix \ref{subsec:image_collection_criteria}, dictating the choice of cultural markers to ensure consistency in the collected data. These criteria aim to reduce misrepresentation of collected cultural data and also ensures variety within each category.



\paragraph{Image Synthesis.}
In preparation for image synthesis, we automatically generate masks of the facial features of each person with the Segment Anything Model (SAM) from Meta~\cite{ravi2024sam2segmentimages}. Then, we conduct image synthesis via inpainting \cite{esser2024scalingrectifiedflowtransformers}, with Stability REST v2 beta API\footnote{\url{https://platform.stability.ai/docs/api-reference}}. Using the original image, generated mask, and target ethnicity as input, the model generates a synthetic image that replaces the human subject with another of a target ethnicity while closely resembling the original. Image synthesis entails replacing the human subject in each original image with another individual with phenotypic traits that align with the prompts used to guide the synthesis, which are provided in Appendix~\ref{subsec:diffusion_model_prompts}. 

\paragraph{Quality Assurance.}
All generated images were vetted by automated flagging and manual human evaluation to minimize artifacts and misrepresentation of a culture. As an automated filter, we use the combination of BRISQUE \cite{6190099} and CLIP similarity \cite{radford2021learningtransferablevisualmodels} as detailed in Appendix~\ref{subsec:automated_filtering}. After filtering out automatically flagged images, each generated image is manually inspected by a human to ensure that the cultural markers remain visually intact, still representing the culture. In cases where artifacts persisted, adjustments were made by manually modifying the mask or further by substituting the original image-mask pair entirely with one that was more suitable for synthesis.

\begin{figure*}[!t]
    \centering
    \captionsetup{justification=centering}
    \includegraphics[width=0.9\textwidth]{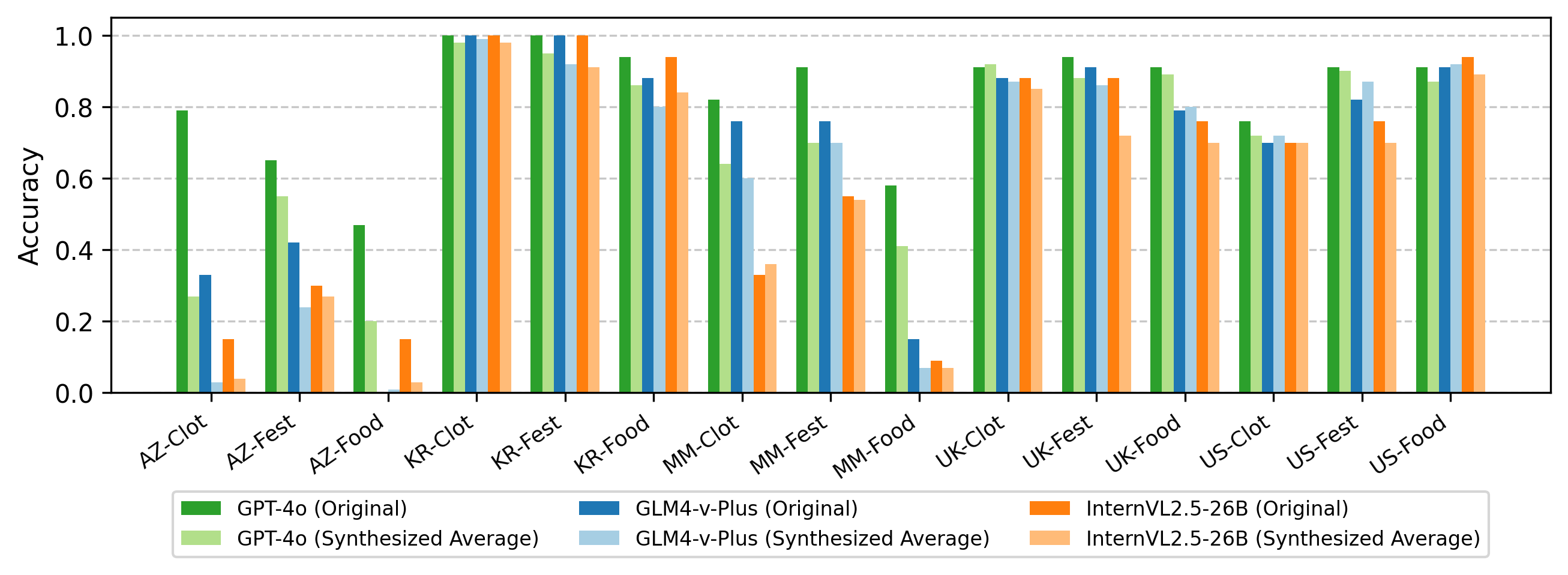}
    \caption{Country Identification accuracy on original images and the average over corresponding synthesized images of four ethnicities (colored in pale) for each country-category pair.}
    \label{fig:cultural_accuracy_barchart_average}
\end{figure*}
\begin{figure*}[!t]
    \centering
    \captionsetup{justification=centering}
    \includegraphics[width=0.9\textwidth]{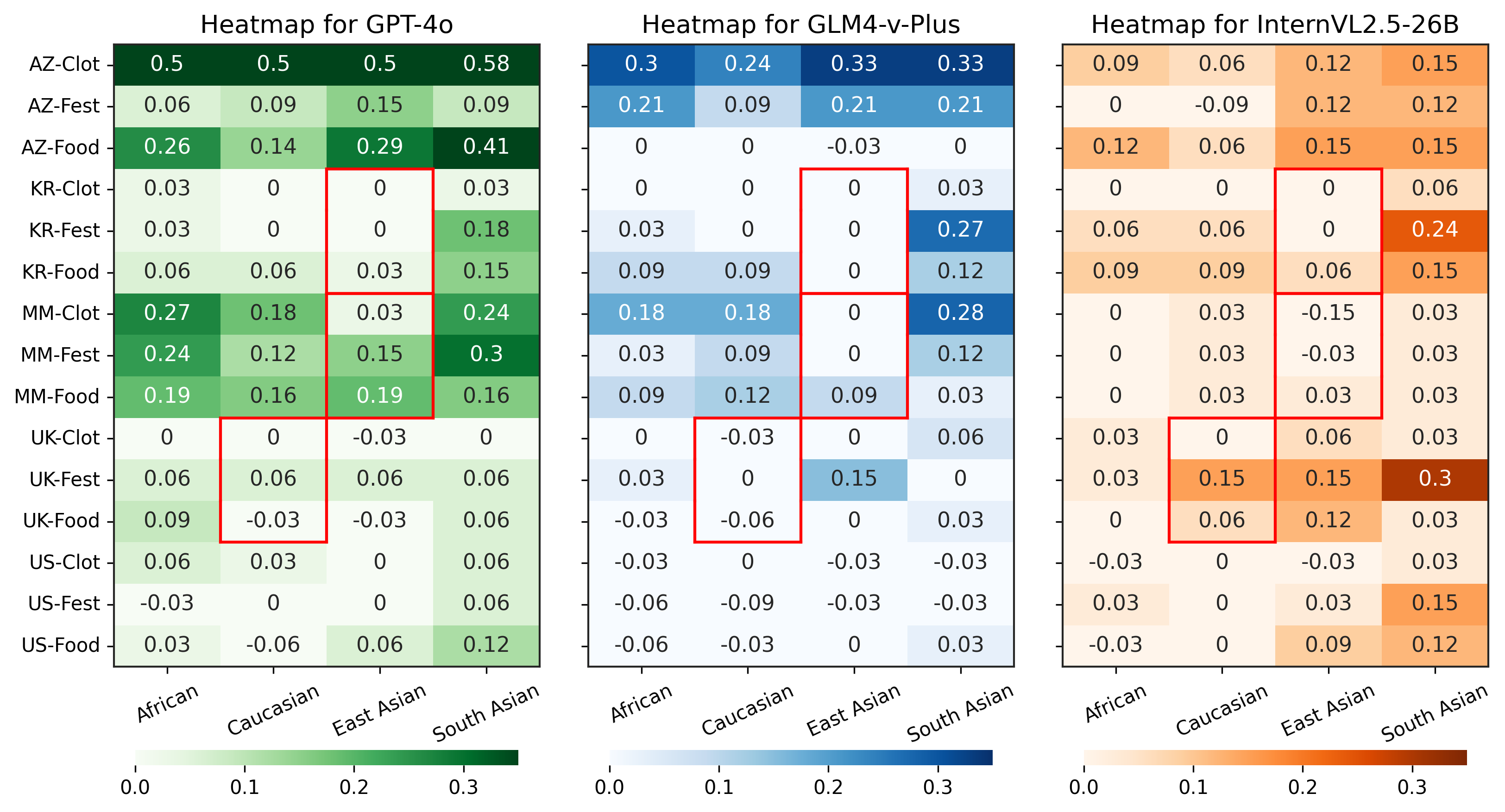}
    \caption{Heatmap of Country Identification accuracy difference. The value in each cell is the difference in Country Identification accuracy between the original and that of synthesized ethnicity. The red boxes highlight the pairs where the synthesized ethnicity by the inpainting model closely resembles a culture demographic.}
    \label{fig:cultural_accuracy_heatmap}
\end{figure*}

Additional details such as the composition of the dataset, the masking procedure, and labeling are described in Appendix~\ref{subsec:dataset_construction_details}.

\section{Evaluating MLLMs with \ours}




We evaluate the cultural bias of MLLMs through two tasks: Country Identification and Cultural Marker Identification. \emph{Country Identification} is the task of identifying the country of origin or culture of a given cultural marker. \emph{Cultural Marker Identification} is the task of identifying the name of a cultural marker. For this task, we focus only on the Food category as foods are the most diverse and distinguishable in terms of their labels. Other categories, Clothes and Festivals, often lack specific names or have only one widely recognized label, presenting a difficulty for even native annotators to identify.

Accuracy is used to quantify the ability of the MLLMs on both tasks. For \emph{Country Identification}, a model's output that did not include the exact ground-truth country or culture verbatim was marked as incorrect. For \textit{Cultural Marker Identification} accuracy, a secondary LLM, GLM-4-Plus\footnote{\url{https://bigmodel.cn/dev/howuse/glm-4}}, is used as an evaluator model, given access to ground-truth labels, to assess whether a response sufficiently identified the food. \footnote{Examples of ground-truth labels are provided in Appendix~\ref{subsec:labels}.}

\paragraph{Does replacing the person in an image with a person of a different ethnicity introduce cultural bias in MLLMs?}
Country Identification accuracy for original images are generally higher than those of synthesized ones as apparent in Figure~\ref{fig:cultural_accuracy_barchart_average} by an average of 7.64\% across all models. However, synthesized ethnicities that closely resemble the demographic of the original culture typically perform better than other ethnicities, achieving an accuracy drop from the original images of just 2.04\% and even occasionally matching or outperforming the original images. For example, images synthesized with East Asian ethnicity demonstrate minimal accuracy drops for Korea and Myanmar compared to alterations into other ethnicities. Similarly, UK images synthesized with Caucasian ethnicity show low sensitivity to alterations, achieving accuracy levels close to those of the originals. The alignment is expected, given that Korea and Myanmar belong to East and Southeast Asia respectively, where visual changes made by the diffusion model for East Asian subjects are minimal. Likewise, since the majority of the UK's population is White British, Caucasian synthesis introduces only trivial visual modifications. In contrast, significant accuracy drops are observed when images are altered to African or South Asian ethnicities, where visual differences are, in general, significant for predominant population of Korea, Myanmar, and the UK. These drops are relative to other synthesized ethnicities within the same country and category.

As can be inferred from Figure \ref{fig:cultural_accuracy_heatmap}, a common trend in robustness among the three MLLMs is that their accuracy is barely affected by ethnicity alteration in the UK and the US with drops in accuracy less than 15\% across all categories and ethnicities (except InternVL-UK-festival case). Also, all three models show significant accuracy drops in Korean Festival and Korean Food for South Asian while being fairly robust in other ethnicities.

Evaluating Myanmar and Azerbaijan, notable sensitivity is observed in GPT-4o and GLM-4v. GPT-4o shows the highest sensitivity (eg. >40\% differences are observed in Azerbaijan) although its absolute accuracy is always higher than the other two models. GLM-4v also exhibits sensitivity but in fewer categories and less intensity than GPT-4o does. Although InternVL is the least sensitive overall, its consistency is partly because of its equally underwhelming accuracy (less than 20\%) across ethnicities in some categories such as AZ-Food, AZ-Clothes and MM-Food.

\begin{figure*}[!h]
    \centering
    \captionsetup{justification=centering}
    \includegraphics[width=0.9\textwidth]{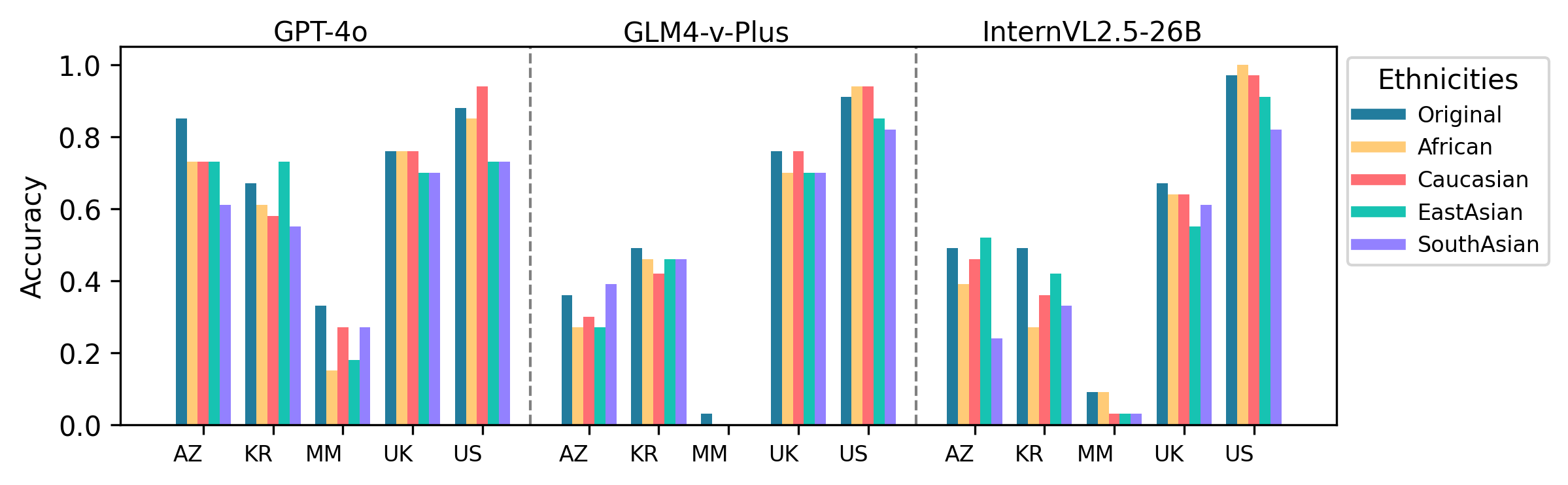}
    \caption{Cultural Marker Identification accuracy evaluated on \textit{Food} images.}
    \label{fig:foodlabel_accuracy_barchart}
\end{figure*}

Cultural Marker Identification accuracies, shown in Figure~\ref{fig:foodlabel_accuracy_barchart}, exhibit similar sensitivity trends to ethnicity changes. Models like GPT-4o and InternVL drop up to 24\% in accuracy for Korean and Azerbaijani food when images are synthesized with South Asian ethnicity. GLM-4v-Plus retains a stable sensitivity across cultures. However, we may still observe for all models that Cultural Marker Identification accuracy values tend to drop for synthesized images, and more so for those that deviate further from the original country's demographic.

\paragraph{How does this bias differ depending on whether the cultural marker belongs to a low-resource or high-resource culture?}
The Country Identification accuracy across different countries serves as a quantitative measure of the cultural resource levels embedded within various MLLMs. Azerbaijan and Myanmar have consistently lower accuracy, compared to the UK, the US and South Korea, which have accuracy within (80\%-100\%) in general. This further validates the current literature \cite{gustafson2023pinpointingobjectrecognitionperformance, pouget2024no} that vision models tend to possess less robust knowledge of underrepresented cultures, highlighting the need to train with more culturally diverse data.

Synthesized images play a crucial role in this analysis by normalizing the distribution of ethnicities across all cultural image sets. This mitigates the unexpected factors introduced by uneven dataset representation, ensuring that accuracy differences are primarily attributed to a model's cultural awareness rather than its familiarity with specific ethnic groups.

\section{Discussion}

The discrepancy between Food Identification accuracy and Country Identification accuracy in cultures like Azerbaijan and Korea, underscores the MLLMs' limitation in contextualizing entities into specific cultural frameworks. In Azerbaijan, Food Identification accuracy was significantly higher than the Country Identification accuracy (AZ-food) across all models. This may be attributed to the fact that many Azerbaijani food images in the dataset are visually similar to dishes from neighboring regions or adjacent cultures, such as the Caucasus, the Middle East, and Turkey. Therefore, it can be easier for MLLMs to identify the generic name of the foods (eg. Plov, Kebabs) than to identify the exact country (eg. Azerbaijan) they associated with when the foods are shared by several cultures, albeit with nuanced visual differences. In such cases, models may recognize the food based on its high-level similarities among its variants from similar cultures, rather than its nuanced distinct cultural attributes. 





\section{Conclusion}



In this study, we introduce \ours{} to evaluate the robustness of multi-modal large language models (MLLMs) and their cultural awareness and bias with cross-cultural perturbed data across five cultures (Azerbaijan, Myanmar, South Korea, the UK, and the US) and three categories (Food, Festivals, and Clothes). Our results reveal that MLLMs disproportionately favor high-resource cultures while exhibiting both uncertainty and inconsistency in their awareness in underrepresented cultures. Our findings highlight the need for more diverse, representative data to improve cultural awareness in AI.




\section*{Limitations and Future Work}

The performance drop for synthesized ethnicities may partly stem from minor inpainting artifacts and subtle distortions of cultural markers still persisted despite the quality control, rather than solely from inherent model biases. Furthermore, data contamination — where original images in pretraining datasets inflate Cultural Identification Accuracy — may cause synthesized images to have lower scores due to their novelty. We also acknowledge that the ethnicity alteration that the inpainting model is prompted for is highly generic. For example, `South Asian' encompasses multitudes of ethnicities. Therefore, the synthesized visual appearance is, by no means, intended to be representative of South Asian, but rather a typical sample generated based on the patterns inherently learned by the inpainting model. 

Since our study is limited to evaluating three MLLMs on five cultures with four generalized ethnic depictions across three categories of cultural markers, future research will expand along these dimensions — the number of MLLMs, the range of cultures, synthesized ethnic depictions, and categories of cultural markers. By increasing the number of original images and employing multiple inpainting tools to average outputs, technical uncertainties can also be mitigated. This will enable more robust, statistically significant conclusions about changes in model-driven cultural awareness and expand the scope of the analysis.

\section*{Ethics Statement}
All studies in this research project were conducted with the approval of  KAIST IRB (KAISTIRB-2025-37). This study evaluates the robustness of MLLMs in cultural awareness to promote transparency, fairness, and inclusivity in artificial intelligence while carefully considering the ethical implications of altering human features such as ethnicity. Our work focuses exclusively on assessing model robustness and biases without endorsement of stereotypes of cultural misrepresentation, using synthetic alterations solely to uncover dependencies on peripheral attributes and foster greater inclusivity in future models. We acknowledge the potential misuse of our methodologies—such as exploiting synthesized data for discriminatory purposes, and thus advocate for the responsible use of the benchmark and related tools within clearly defined ethical and scientific boundaries. 

We acknowledge that our positionality as researchers—including our cultural and social backgrounds—may pose an influence on our approach to assessing bias within MLLMs. We remain committed to transparency within our methodology and strive for objectivity.
Additionally, we understand the risks involved in the reinforcement of stereotypes that may arise during the image synthesis stage. To minimize this, our research emphasizes that no culturally connected elements were synthesized, with models instead focused solely on altering the ethnic aspects of each image. Furthermore, the focus of our research is conducted in an effort to quantify the potential reliance of MLLMs on stereotypical markers in an effort to reduce such biases in future models.

\section*{Acknowledgements}
This research was supported by the MSIT(Ministry of Science, ICT), Korea, under the National Program for Excellence in SW), supervised by the IITP(Institute of Information \& communications Technology Planing \& Evaluation) in 2024 (2022-0-01092).

\bibliography{anthology,myrefs}

\clearpage
\appendix
\section*{Appendix}

\section{Details of Dataset Construction}
\label{subsec:dataset_construction_details}
\begin{figure*}[!ht]
    \centering
    \captionsetup{justification=centering}
    \includegraphics[width=\textwidth]{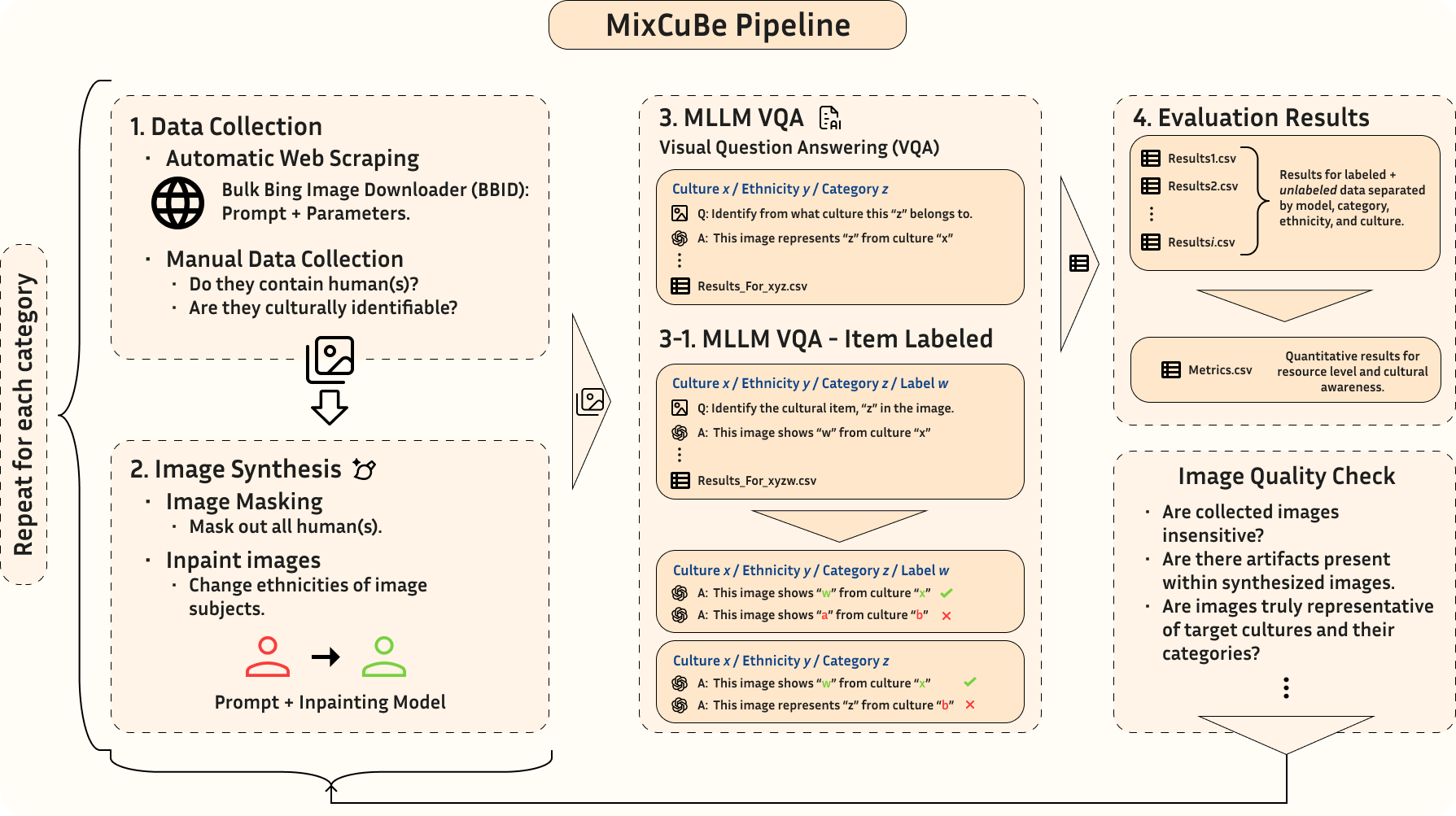}
    \caption{The overall pipeline of the construction of \ours{} and the evaluation of cultural awareness}
    \label{fig:overall_pipeline}
\end{figure*}

Our dataset is publicly accessible at \url{https://huggingface.co/datasets/kyawyethu/MixCuBe}, and it includes original images, synthesized images, and masks. Additionally, a list of labels for food item names have been provided within the dataset. The pipeline illustration for the data set construction is provided in ~Figure \ref{fig:overall_pipeline}.
\subsection{Reference for Country ISO codes and Category Abbreviations}
Throughout the paper, we use the two-letter ISO codes
for each country and four-letter abbreviations for each category of cultural marker as follows.

\begin{table}[!h]\centering
\begin{tabular}{@{}ll@{}}
\toprule
\textbf{Country/Culture} & \textbf{Abbreviation} \\ \midrule
Azerbaijan & AZ \\
South Korea & KR \\
Myanmar & MM \\
United Kingdom & UK \\
United States & US \\ 
\midrule
\textbf{Category} & \textbf{Abbreviation} \\
\midrule
Clothes & Clot \\
Festival & Fest \\
Food & Food \\
\bottomrule
\end{tabular}
\caption{Reference for country ISO codes and abbreviations of categories}
\end{table}

\subsection{Composition of the Dataset}
\label{subsec:dataset_composition}
\ours{} consists of 2.5k labeled images spanning 

- 5 cultures: \textit{Azerbaijan, South Korea, Myanmar, the United Kingdom, and the United States}

- 3 identifying categories of cultural
markers: \textit{Food, Festival, and Clothing}

- 4 Synthesized Ethnicities: \textit{African, Caucasian, East Asian and South Asian}t

For each category of each culture, we collected 33 original images, which were then synthesized by inpainting to generate four sets of synthesized images. The total data is composed of 2475 images.

\begin{table}[h!]\centering
\scriptsize
\begin{tabular}{@{}lllll@{}}
\toprule
\makecell{\textbf{Country} \\/ \textbf{Culture}}         & \makecell{\textbf{Original} \\ \textbf{Images}} & \makecell{\textbf{Synthesized} \\ \textbf{Images}} & \textbf{Categories} & \makecell{\textbf{Total} \\ \textbf{Images}} \\ \midrule
AZ       & 33  & $33 \times 4 $  & $\times 3$  & $33 \times 5 \times 3 = 495$ \\  
KR      & 33  & $33 \times 4 $  & $\times 3$  & $33 \times 5 \times 3 = 495$ \\  
MM        & 33  & $33 \times 4 $  & $\times 3$  & $33 \times 5 \times 3 = 495$ \\  
UK           & 33  & $33 \times 4 $  & $\times 3$  & $33 \times 5 \times 3 = 495$ \\  
US         & 33  & $33 \times 4 $  & $\times 3$  & $33 \times 5 \times 3 = 495$ \\  
\bottomrule
\end{tabular}

\caption{The Composition of the Dataset}
\label{tab:dataset_information_copy}
\end{table}

\subsection{Image Collection Criteria}
\label{subsec:image_collection_criteria}
\begin{enumerate}
    \item Annotators for each set of images within a culture must be native for that culture.
    \item The cultural marker(s) in each image must be easily identifiable by native annotators. (The cultural marker should be both visually clear and popular enough among their culture.)
    \item When choosing images, cultural overlap must be minimized (e.g. American pizza is avoided because while its nuances are specific to the US, pizza, in general, is a very common food eaten worldwide.)
    \item The number of types in a category must be at least a fourth of the total number of images in that category. (e.g. 33 food images → 8 different types of food.)
\end{enumerate}
Additionally, we aim to ensure that the ethnic composition of people in the images for each country in the dataset reflects the demographic makeup of that country.

\subsection{Masking Procedure}
\label{subsec:masking_procedure}
Masking was primarily automated using YuNet\footnote{\url{https://docs.opencv.org/4.x/df/d20/classcv_1_1FaceDetectorYN.html}} from OpenCV to automatically obtain the coordinates of faces as an input to Segment Anything Model (SAM).

For images containing multiple individuals, we limit the number of faces to detect by filtering out the faces with confidence score lower than 0.65 and selecting at most three faces with top confidence if any. This is to ensure only the central and most prominently visible humans are masked, as current inpainting models tend to show a degredation in performance when required to modify multiple subjects simultaneously. Gaussian blur is applied after mask generation to feather the edges of our mask generally helps provide a better inpainting result.

\subsection{Labels}
\label{subsec:labels}
 Multiple acceptable labels for Azerbaijan, the UK and the US were considered as follows to eliminate false negatives in country identification.
 \begin{tcolorbox}[colback=gray!20, colframe=gray!80, boxrule=0.1mm, arc=1mm, fontupper=\small]
 Azerbaijan: \textit{"Azerbaijan, Azerbaijani, Azeri"}\\
 
 UK: \textit{"UK, United Kingdom, Scotland, Britain, British, Irish, Wales, England, English"}\\

 US: \textit{"USA, US, the United States of America, the United States, Hawaii, American"}
\end{tcolorbox}

Likewise, we pre-defined multiple acceptable ground-truth labels for food of each \textit{Food} image to aid the evaluator model in its assessment. A label can be either unique or shared across multiple images. One food label for each country is provided as examples in the following.

\begin{tcolorbox}[colback=gray!20, colframe=gray!80, boxrule=0.1mm, arc=1mm, fontupper=\small]
Azerbaijan: "Azerbaijanian Baklava"\\

Korea: "Jjajangmyeon, Kimchi, Ramyeon, Black Bean Noodles"\\

Myanmar: "Laphet Thoke, Tea leaf salad"\\

UK: "Cottage Pie, Shepherds Pie, Shepherd"\\

US: "Grilled cheese, Toastie"
 \end{tcolorbox}

\subsection{Automated Filtering for Synthesized Images}
\label{subsec:automated_filtering}
The automated flagging technique we employed partially follows the procedure of image saliency check from \cite{pal2024semitruthslargescaledatasetaiaugmented}. BRISQUE, a reference-free metric that quantifies the perceptual quality of an image, is used to detect images with low structural integrity, indicated by a high BRISQUE score. To ensure that the inpainting model performs enough augmentations on the original while retaining certain resemblance to the original, images beyond the defined range of CLIP similarity are flagged. A synthesized image with low CLIP similarity cannot impose a sufficient visual challenge on MLLMs while extreme visual divergence from the original may potentially distort the cultural marker that the model is supposed to identify. Therefore, images with either BRISQUE score of greater than 80 or CLIP similarity outside the range (65 - 98) are discarded. 


\section{Prompts}
\subsection{Prompts for Diffusion Inpaiting Model}
\label{subsec:diffusion_model_prompts}
For inpainting of images in batch using Stable diffusion, we use a general prompt for each category as follows.
\begin{tcolorbox}[colback=gray!20, colframe=gray!80, boxrule=0.1mm, arc=1mm, fontupper=\small]

Clothes: \textit{"\{Ethnicity\} person(s) in clothes"}
\\

Food: \textit{"\{Ethnicity\} person(s) with food"}
\\

Festival: \textit{"\{Ethnicity\} person(s) at an occasion"}
\end{tcolorbox}
The placeholder, \textit{\{Ethnicity\}}, is one of African, Caucasian, East Asian, South Asian.

We prompt the model again with a tailored prompt for each image having undesirable result from the initial batch inpainting. Some examples are

- \textit{"South Asian men playing a flute"}

- \textit{"Caucasian ladies performing a dance"}

- \textit{"An East Asian couple in Myanmar traditional clothes"}

\subsection{Prompts for Evaulating MLLMs}
For evaluating MLLMs, we use a dedicated prompt for each category as follows.
\begin{tcolorbox}[colback=gray!20, colframe=gray!80, boxrule=0.1mm, arc=1mm, fontupper=\small]

Clothes: \textit{"Which country is the clothing in the photo the most associated with? Which visual cues did you use to determine it?"}
\\

Food: \textit{"Which country is the food in the photo the most associated with? Which food is it?"}
\\

Festival: \textit{"Which country is the celebration/cultural activity/festival in the photo the most associated with? Which visual cues did you use to determine it?""}
\end{tcolorbox}

\subsection{Prompt for Evaluator LLM of Responses from Food Images}
\label{subsec:prompt_evaluator_LLM}
For determining whether or not a response from a MLLM sufficiently and correctly identifies the food in an image, we used GLM-4-Plus as an evaluator LLM with the following one-shot prompt.

\begin{tcolorbox}[colback=gray!20, colframe=gray!80, boxrule=0.1mm, arc=1mm, fontupper=\small]
        You'll be provided with a label and a response by a multi-modal LLM that identifies the name of the food in an image. Determine whether the food name contained in the response can be considered as correct given the ground-truth label. under label. Consider it as correct ('Yes') if the names of the food refer to the same food semantically either in native language or in English. Otherwise, 'No'. \\
        
        - Emphasize on the name instead of the description. \\ 
        
        - The names do not need to match exactly. \\ 
        
        - If the name provided is wrong, it's 'No' even if the description is close. For example, if the label is \textit{"Dote Htoe,Wat Thar Dote Htoe,pork offal skewers,pork skwers"} and the response is \textit{"This food is mostly associated with Myanmar and is called 'E Kya Kway' or 'Inn Kyaik Kyaw'. It's a popular street food featuring skewers, often with a variety of meats and offal, cooked in a boiling broth."}, the answer should be 'No' since the name is completely wrong and the description does not include 'pork'. \\  
        
        - Answer only in 'Yes' or 'No'.
\end{tcolorbox}

\section{Experimental Settings}
\subsection{Models}

We evaluated the cultural awareness of three MLLMs — GPT-4o (\texttt{gpt-4o-2024-08-06})\footnote{\url{https://platform.openai.com/docs/models}} by OpenAI, GLM-4v (\texttt{glm-4v-plus})\footnote{\url{https://bigmodel.cn/dev/howuse/glm-4v}} by ZhipuAI, and InternVL2.5 (\texttt{InternVL2.5-26B-AWQ})\footnote{\url{https://huggingface.co/OpenGVLab/InternVL2_5-26B-AWQ}} by OpenGVLab. 


\subsection{Hyperparameters}
The following table provides the values of some key hyperparameters used in the experiments.
\begin{table}[h!]\centering
\small
\begin{tabular}{@{}lll@{}}
\toprule
\textbf{Model}         & \textbf{Hyperparameter}       & \textbf{Value}          \\ \midrule
Stable Diffusion       & Diffusion steps               & 60                                     \\
                       & Guidance scale                & 7.0                     \\ \cmidrule(l){1-3}
GPT-4o, GLM-4v,        & Maximum Token & 120 \\
InternVL2.5            & Temperature & 0.3  \\ 
                       & Top-p & 0.6 \\
                       \cmidrule(l){1-3}
GLM-4v                  & Maximum Token & 10 \\
                        & Temperature & 0.2  \\ 
                       & Top-p & 0.5                     \\ 
                       \bottomrule
\end{tabular}
\caption{Hyperparameters used in the experiments.}
\label{tab:hyperparameters}
\end{table}

\section{MLLM Evaluation Results}

\begin{figure*}[!ht]
        \begin{threeparttable}
\small
\setlength{\tabcolsep}{2pt} 
\begin{tabular}{lrrrrrrrrrrr}\toprule

\multirow{2}{*}{\makecell{\textbf{Country}/ \\ \textbf{Culture}}} 
& \multirow{2}{*}{\textbf{Ethnicity}} 
& \multicolumn{3}{c}{\textbf{Clothes}} 
& \multicolumn{3}{c}{\textbf{Food}} 
& \multicolumn{3}{c}{\textbf{Festival}} \\\cmidrule{3-11}

& 
& \makecell{\textbf{GPT-4o}} 
& \makecell{\textbf{GLM-4v} \\ {\textbf{-Plus}}} 
& \makecell{\textbf{InternVL} \\ \textbf{2.5-26B}} 
& \makecell{\textbf{GPT-4o}} 
& \makecell{\textbf{GLM-4v} \\ {\textbf{-Plus}}} 
& \makecell{\textbf{InternVL} \\ \textbf{2.5-26B}} 
& \makecell{\textbf{GPT-4o}} 
& \makecell{\textbf{GLM-4v} \\ {\textbf{-Plus}}} 
& \makecell{\textbf{InternVL} \\ \textbf{2.5-26B}} \\ \midrule

\multirow{6}{*}{Azerbaijan} &\cellcolor[HTML]{c0edda}Original &\cellcolor[HTML]{c0edda}0.79 &\cellcolor[HTML]{c0edda}0.47 &\cellcolor[HTML]{c0edda}0.65 &\cellcolor[HTML]{c0edda}0.33 &\cellcolor[HTML]{c0edda}0.00 &\cellcolor[HTML]{c0edda}0.42 &\cellcolor[HTML]{c0edda}0.15 &\cellcolor[HTML]{c0edda}0.15 &\cellcolor[HTML]{c0edda}0.30 \\
&African &0.29 &0.21 &0.59 &0.03 &0.00 &0.21 &0.06 &0.03 &0.30 \\
&Caucasian &0.29 &0.33 &0.56 &0.09 &0.00 &0.33 &0.09 &0.09 &0.39 \\
&East Asian &0.29 &0.18 &0.50 &0.00 &0.03 &0.21 &0.03 &0.00 &0.18 \\
&South Asian &0.21 &0.06 &0.56 &0.00 &0.00 &0.21 &0.00 &0.00 &0.18 \\
&\cellcolor[HTML]{c0edda}Average (Synthesized) &\cellcolor[HTML]{c0edda}0.27 &\cellcolor[HTML]{c0edda}0.20 &\cellcolor[HTML]{c0edda}0.55 &\cellcolor[HTML]{c0edda}0.03 &\cellcolor[HTML]{c0edda}0.01 &\cellcolor[HTML]{c0edda}0.24 &\cellcolor[HTML]{c0edda}0.04 &\cellcolor[HTML]{c0edda}0.03 &\cellcolor[HTML]{c0edda}0.27 \\
\midrule
\multirow{6}{*}{Korea} &\cellcolor[HTML]{c0edda}Original &\cellcolor[HTML]{c0edda}1.00 &\cellcolor[HTML]{c0edda}0.94 &\cellcolor[HTML]{c0edda}1.00 &\cellcolor[HTML]{c0edda}1.00 &\cellcolor[HTML]{c0edda}0.88 &\cellcolor[HTML]{c0edda}1.00 &\cellcolor[HTML]{c0edda}1.00 &\cellcolor[HTML]{c0edda}0.94 &\cellcolor[HTML]{c0edda}1.00 \\
&African &0.97 &0.88 &0.97 &1.00 &0.79 &0.97 &1.00 &0.85 &0.94 \\
&Caucasian &1.00 &0.88 &1.00 &1.00 &0.79 &1.00 &1.00 &0.85 &0.94 \\
&East Asian &1.00 &0.91 &1.00 &1.00 &0.88 &1.00 &1.00 &0.88 &1.00 \\
&South Asian &0.97 &0.79 &0.82 &0.97 &0.76 &0.73 &0.94 &0.79 &0.76 \\
&\cellcolor[HTML]{c0edda}Average (Synthesized) &\cellcolor[HTML]{c0edda}0.98 &\cellcolor[HTML]{c0edda}0.86 &\cellcolor[HTML]{c0edda}0.95 &\cellcolor[HTML]{c0edda}0.99 &\cellcolor[HTML]{c0edda}0.80 &\cellcolor[HTML]{c0edda}0.92 &\cellcolor[HTML]{c0edda}0.98 &\cellcolor[HTML]{c0edda}0.84 &\cellcolor[HTML]{c0edda}0.91 \\
\midrule
\multirow{6}{*}{Myanmar} &\cellcolor[HTML]{c0edda}Original &\cellcolor[HTML]{c0edda}0.82 &\cellcolor[HTML]{c0edda}0.58 &\cellcolor[HTML]{c0edda}0.91 &\cellcolor[HTML]{c0edda}0.76 &\cellcolor[HTML]{c0edda}0.15 &\cellcolor[HTML]{c0edda}0.76 &\cellcolor[HTML]{c0edda}0.33 &\cellcolor[HTML]{c0edda}0.09 &\cellcolor[HTML]{c0edda}0.55 \\
&African &0.55 &0.39 &0.67 &0.58 &0.06 &0.73 &0.33 &0.09 &0.55 \\
&Caucasian &0.64 &0.42 &0.79 &0.58 &0.03 &0.67 &0.30 &0.06 &0.52 \\
&East Asian &0.79 &0.39 &0.76 &0.76 &0.06 &0.76 &0.48 &0.06 &0.58 \\
&South Asian &0.58 &0.42 &0.61 &0.48 &0.12 &0.64 &0.30 &0.06 &0.52 \\
&\cellcolor[HTML]{c0edda}Average (Synthesized) &\cellcolor[HTML]{c0edda}0.64 &\cellcolor[HTML]{c0edda}0.41 &\cellcolor[HTML]{c0edda}0.70 &\cellcolor[HTML]{c0edda}0.60 &\cellcolor[HTML]{c0edda}0.07 &\cellcolor[HTML]{c0edda}0.70 &\cellcolor[HTML]{c0edda}0.36 &\cellcolor[HTML]{c0edda}0.07 &\cellcolor[HTML]{c0edda}0.54 \\
\midrule
\multirow{6}{*}{UK} &\cellcolor[HTML]{c0edda}Original &\cellcolor[HTML]{c0edda}0.91 &\cellcolor[HTML]{c0edda}0.91 &\cellcolor[HTML]{c0edda}0.94 &\cellcolor[HTML]{c0edda}0.88 &\cellcolor[HTML]{c0edda}0.79 &\cellcolor[HTML]{c0edda}0.91 &\cellcolor[HTML]{c0edda}0.88 &\cellcolor[HTML]{c0edda}0.76 &\cellcolor[HTML]{c0edda}0.88 \\
&African &0.91 &0.82 &0.88 &0.88 &0.82 &0.88 &0.85 &0.76 &0.85 \\
&Caucasian &0.91 &0.94 &0.88 &0.91 &0.85 &0.91 &0.88 &0.70 &0.73 \\
&East Asian &0.94 &0.94 &0.88 &0.88 &0.79 &0.76 &0.82 &0.64 &0.73 \\
&South Asian &0.91 &0.85 &0.88 &0.82 &0.76 &0.91 &0.85 &0.73 &0.58 \\
&\cellcolor[HTML]{c0edda}Average (Synthesized) &\cellcolor[HTML]{c0edda}0.92 &\cellcolor[HTML]{c0edda}0.89 &\cellcolor[HTML]{c0edda}0.88 &\cellcolor[HTML]{c0edda}0.87 &\cellcolor[HTML]{c0edda}0.80 &\cellcolor[HTML]{c0edda}0.86 &\cellcolor[HTML]{c0edda}0.85 &\cellcolor[HTML]{c0edda}0.70 &\cellcolor[HTML]{c0edda}0.72 \\
\midrule
\multirow{6}{*}{US} &\cellcolor[HTML]{c0edda}Original &\cellcolor[HTML]{c0edda}0.76 &\cellcolor[HTML]{c0edda}0.91 &\cellcolor[HTML]{c0edda}0.91 &\cellcolor[HTML]{c0edda}0.70 &\cellcolor[HTML]{c0edda}0.91 &\cellcolor[HTML]{c0edda}0.82 &\cellcolor[HTML]{c0edda}0.70 &\cellcolor[HTML]{c0edda}0.94 &\cellcolor[HTML]{c0edda}0.76 \\
&African &0.70 &0.88 &0.94 &0.73 &0.97 &0.88 &0.73 &0.97 &0.73 \\
&Caucasian &0.73 &0.97 &0.91 &0.70 &0.94 &0.91 &0.70 &0.94 &0.76 \\
&East Asian &0.76 &0.85 &0.91 &0.73 &0.91 &0.85 &0.73 &0.85 &0.73 \\
&South Asian &0.70 &0.79 &0.85 &0.73 &0.88 &0.85 &0.67 &0.82 &0.61 \\
&\cellcolor[HTML]{c0edda}Average (Synthesized) &\cellcolor[HTML]{c0edda}0.72 &\cellcolor[HTML]{c0edda}0.87 &\cellcolor[HTML]{c0edda}0.90 &\cellcolor[HTML]{c0edda}0.72 &\cellcolor[HTML]{c0edda}0.92 &\cellcolor[HTML]{c0edda}0.87 &\cellcolor[HTML]{c0edda}0.70 &\cellcolor[HTML]{c0edda}0.89 &\cellcolor[HTML]{c0edda}0.70 \\
\bottomrule
\end{tabular}
\begin{tablenotes}
    \small
    \item Note: Each cell represents accuracy percentage calculated out of 33 images except cells in the row of Average (Synthesized).
\end{tablenotes}
\end{threeparttable}
        \caption{Country Identification Accuracy Data}
        \label{tab:cultural_accuracy_table}
\end{figure*}
\begin{figure*}[htbp]
    \centering
    \begin{subfigure}{0.50\textwidth}
        \begin{threeparttable}
\small
\begin{tabular}{lrrrrr}\toprule
\makecell{\textbf{Country}/ \\ \textbf{Culture}} &\textbf{Ethnicity} &\textbf{GPT4-o}  & \makecell{\textbf{GLM4-v} \\ \textbf{-Plus}} 
& \makecell{\textbf{InternVL} \\ \textbf{2.5-26B}} \\ \midrule

\multirow{6}{*}{Azerbaijan} &\cellcolor[HTML]{c0edda}Original &\cellcolor[HTML]{c0edda}0.85 &\cellcolor[HTML]{c0edda}0.36 &\cellcolor[HTML]{c0edda}0.49 \\
&African &0.73 &0.27 &0.39 \\
&Caucasian &0.73 &0.30 &0.46 \\
&East Asian &0.73 &0.27 &0.52 \\
&South Asian &0.61 &0.39 &0.24 \\
&\cellcolor[HTML]{c0edda}\makecell{Average \\ (Synthesized)} &\cellcolor[HTML]{c0edda}0.70 &\cellcolor[HTML]{c0edda}0.31 &\cellcolor[HTML]{c0edda}0.40 \\
\midrule
\multirow{6}{*}{Korea} &\cellcolor[HTML]{c0edda}Original &\cellcolor[HTML]{c0edda}0.67 &\cellcolor[HTML]{c0edda}0.49 &\cellcolor[HTML]{c0edda}0.49 \\
&African &0.61 &0.46 &0.27 \\
&Caucasian &0.58 &0.42 &0.36 \\
&East Asian &0.73 &0.46 &0.42 \\
&South Asian &0.55 &0.46 &0.33 \\
&\cellcolor[HTML]{c0edda}\makecell{Average \\ (Synthesized)} &\cellcolor[HTML]{c0edda}0.61 &\cellcolor[HTML]{c0edda}0.45 &\cellcolor[HTML]{c0edda}0.35 \\
\midrule
\multirow{6}{*}{Myanmar} &\cellcolor[HTML]{c0edda}Original &\cellcolor[HTML]{c0edda}0.33 &\cellcolor[HTML]{c0edda}0.03 &\cellcolor[HTML]{c0edda}0.09 \\
&African &0.15 &0.00 &0.09 \\
&Caucasian &0.27 &0.00 &0.03 \\
&East Asian &0.18 &0.00 &0.03 \\
&South Asian &0.27 &0.00 &0.03 \\
&\cellcolor[HTML]{c0edda}\makecell{Average \\ (Synthesized)} &\cellcolor[HTML]{c0edda}0.22 &\cellcolor[HTML]{c0edda}0.00 &\cellcolor[HTML]{c0edda}0.05 \\
\midrule
\multirow{6}{*}{UK} &\cellcolor[HTML]{c0edda}Original &\cellcolor[HTML]{c0edda}0.76 &\cellcolor[HTML]{c0edda}0.76 &\cellcolor[HTML]{c0edda}0.67 \\
&African &0.76 &0.70 &0.64 \\
&Caucasian &0.76 &0.76 &0.64 \\
&East Asian &0.70 &0.70 &0.55 \\
&South Asian &0.70 &0.70 &0.61 \\
&\cellcolor[HTML]{c0edda}\makecell{Average \\ (Synthesized)} &\cellcolor[HTML]{c0edda}0.73 &\cellcolor[HTML]{c0edda}0.71 &\cellcolor[HTML]{c0edda}0.61 \\
\midrule
\multirow{6}{*}{US} &\cellcolor[HTML]{c0edda}Original &\cellcolor[HTML]{c0edda}0.88 &\cellcolor[HTML]{c0edda}0.91 &\cellcolor[HTML]{c0edda}0.97 \\
&African &0.85 &0.94 &1.00 \\
&Caucasian &0.94 &0.94 &0.97 \\
&East Asian &0.73 &0.85 &0.91 \\
&South Asian &0.73 &0.82 &0.82 \\
&\cellcolor[HTML]{c0edda}\makecell{Average \\ (Synthesized)} &\cellcolor[HTML]{c0edda}0.81 &\cellcolor[HTML]{c0edda}0.89 &\cellcolor[HTML]{c0edda}0.92 \\

\bottomrule
\end{tabular}
\begin{tablenotes}
    \small
    \item Note: Each cell represents accuracy percentage calculated out of 33 images except cells in the row of Average (Synthesized).
\end{tablenotes}
\end{threeparttable}

        \caption{}
        \label{tab:food_accuracy_table}
    \end{subfigure}
    \hfill
    \begin{subfigure}{0.44\textwidth}
        \includegraphics[width=\linewidth]{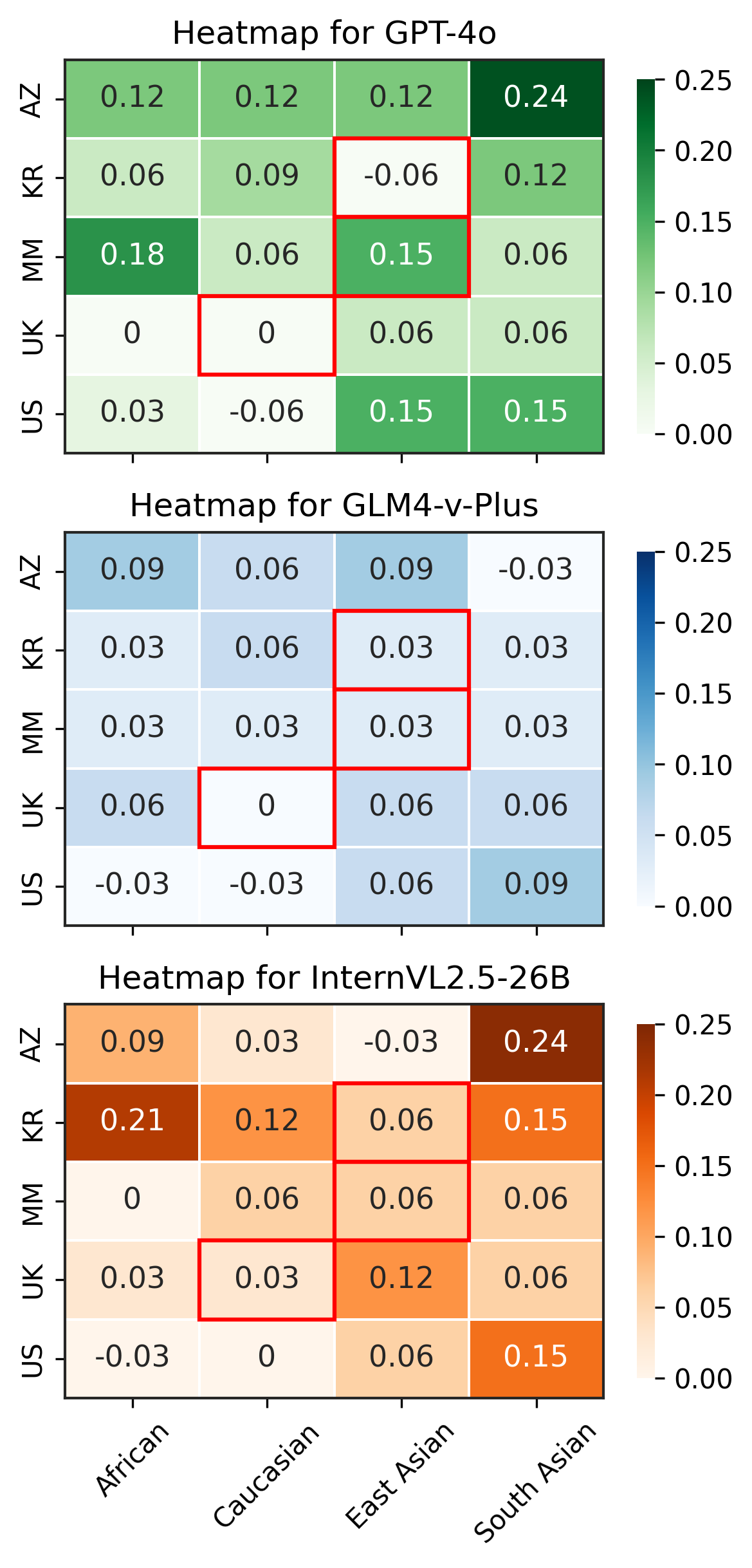}
        \caption{}
        \label{fig:foodlabel_accuracy_heatmap}
    \end{subfigure}

    \caption{(a) Cultural Marker Identification Accuracy Data(b) Cultural Marker Identification Accuracy Difference Heatmap. The value in each cell is the difference in Cultural Marker Identification Accuracy between the original and that of synthesized ethnicity. The red boxes highlight the pairs where the synthesized ethnicity by the inpainting model closely resemble to a demographic of the culture.}
    \label{fig:llm_performance}
\end{figure*}

Table \ref{tab:cultural_accuracy_table} and Figure \ref{fig:cultural_accuracy_heatmap} collectively show the results of country identification, presenting, in each cell, the absolute accuracy and the difference in accuracy with respect to that of the original respectively. Likewise, Table \ref{tab:food_accuracy_table} and Figure \ref{fig:foodlabel_accuracy_heatmap} display the results of Cultural Marker Identification.

\end{document}